\documentclass[conference]{IEEEtran}
\usepackage{booktabs,graphicx,amsmath,amssymb,url,xcolor,algorithm}
\usepackage[noend]{algpseudocode}
\usepackage[hidelinks]{hyperref}
\newcommand{\best}[1]{\textbf{#1}}
\title{OVMAN: A Task and Benchmark for Open-Vocabulary Motion-Aware Navigation}
\author{Dibyendu Ghosh}

\begin{document}
\maketitle

\begin{abstract}
Homes change between a robot's visits. Navigation benchmarks pose their goals in the world the
agent currently sees, and the two-visit benchmarks that exist score recall or rearrangement rather
than navigation. None of them can express \emph{go to the chair that was moved} or \emph{go to
where the vase used to be}. OVMAN is a task in which an agent tours a scene, returns after a
scripted change, and must navigate to a goal specified by the change itself. Two of its six change
relations answer with a place an object has left, where nothing remains to be detected. We release
219 two-visit episodes, each certified solvable by an oracle agent that completes it three times.
Two released systems fail as predicted. A zero-shot object-goal navigator reaches the
change-defined target in 12.3\% of episodes and almost never reaches a vacated location,
0.026 on \textsc{former} and 0.050 on \textsc{removed}. When a self-maintaining
open-vocabulary map is read as two visits rather than as one maintained map, overall success rises
(0.396 to 0.479 on our mapping) but the past-position relations specifically do not recover,
on our maps or on a released self-maintaining one; keeping a map current is therefore not the whole
obstacle. A simple two-visit reference agent reaches
45.2\% when navigating, against an embodied oracle of 99.5\%. An error decomposition places the
remaining difficulty in carrying an instance identity across visits rather than in naming it or in
selecting the answer once positions are known.
\end{abstract}

\section{Introduction}\label{sec:intro}
A robot that lives in a home returns to the same rooms every day, and between its visits the
world moves: a chair is pushed to the table, a vase is carried away, a box appears in the
hallway. ``Put back the chair that was moved'' and ``where did the vase go?'' are ordinary
requests, and both refer to a state of the world that no longer holds.

\textbf{The pieces exist separately; the combination does not.} Visual Room
Rearrangement (RoomR)~\cite{weihs2021roomr} gives an agent a walkthrough, changes objects while it is
absent, and asks it to restore them, but its goal is implicit and geometric and never names an
object. Episodic Memory Question Answering (EMQA)~\cite{datta2022emqa} localises where an object was
last seen as a heatmap over a floorplan, from a closed set of categories and without navigating. FindingDory~\cite{yadav2025findingdory} conditions navigation
goals on the agent's own past interactions rather than on a change. 3D Semantic
MapNet (3D-SMNet)~\cite{cartillier2024smnet} re-identifies objects that moved, vanished or appeared between
two tours and stops at perception. DynaMem~\cite{liu2024dynamem}, OpenIN~\cite{openin2025} and
OpenObject-NAV (OO-NAV)~\cite{tang2025openobjectnav} navigate to objects whose position changes, and
OpenIN and OO-NAV resolve a specific instance rather than a category. All three route to where the
instance is \emph{now}; none poses a goal at a place an object has left.

The canonical navigation tasks each fixed one axis: ObjectNav~\cite{batra2020objectnav} the
semantic goal and the stop-within-$1$\,m convention we inherit, HM3D-OVON~\cite{yokoyama2024hm3dovon}
the open vocabulary, VLN~\cite{anderson2018vln} and VLN-CE~\cite{krantz2020vlnce} the procedural
instruction, Habitat~3.0~\cite{puig2023habitat3} humans who move during the episode, and
OVMM~\cite{yenamandra2023homerobot} the compositional goal with per-stage success, a diagnostic
style we adopt. In all of them the agent meets the world once and the goal is present when the
episode begins.

\textbf{Current-state maps supply reference but not history.} VLMaps~\cite{huang2023vlmaps},
ConceptGraphs~\cite{gu2024conceptgraphs} and VLFM~\cite{yokoyama2024vlfm} say \emph{what is
where} within one session; they do not retain the earlier state a past-tense query is about. Systems that do handle change keep the
map current rather than the past: DualMap~\cite{jiang2025dualmap} and OpenIN~\cite{openin2025}
update object state online, so the superseded position is no longer represented in the map a
query is answered from. Change detection~\cite{objectscanmove2022,ndfchange2022,3dvsg2022} compares scans, online or
offline, and ends at detection or at a planner that seeks changes, with no language goal; maps
that attach a motion attribute to each object, including Vision--Language--Motion
Maps~\cite{ghosh2026vlmm}, classify what moves from online observations but pose no goal, and
ReMEmbR~\cite{anwar2024remembr} answers long-horizon questions about a robot's own history yet
navigates only to what is present.
Closest in spirit, SuperMap~\cite{zhao2026supermap} keeps instance history in an online 4D scene
graph and detects appearance, disappearance and relocation, but it is scored as change-detection
recall on six objects and its language navigation is demonstrated on relations between present
objects; a goal defined by the change, or at a vacated location, is not posed.

This paper makes two contributions.
\begin{itemize}\setlength{\itemsep}{1pt}\setlength{\parskip}{0pt}
  \item \textbf{A task: navigation to a change.} The goal referent is selected by a change relation
        between two visits, not by a category or a route, and the answer can be a location where
        nothing is. The six
        relations are predicates on two world states, Eq.~\eqref{eq:relations}; the success test
        covers vacated locations, Eq.~\eqref{eq:success}; and an anti-shortcut rule keeps category
        recognition at chance (Section~\ref{sec:task}).
  \item \textbf{A benchmark with certified solvability.} 219 two-visit episodes in 20 scenes, each
        kept only after an oracle agent has completed it three times and its recorded observations
        have been checked against the evaluated world (Algorithm~\ref{alg:gen}). It is released with the scoring
        protocol, adapters that run published systems from their own code, and a reference agent
        with every constant stated (Sections~\ref{sec:validity} and~\ref{sec:agent}).
\end{itemize}

Table~\ref{tab:position} places OVMAN against the closest prior work along the four axes the task
needs. No individual column is new, and no row other than ours occupies all four: a goal whose
\emph{referent is selected by a cross-visit change relation}, expressed in open vocabulary, with
answers that may be locations where nothing is.

\begin{table}[t]\centering\footnotesize
\caption{OVMAN against the closest prior work. No individual column is new; no other row
occupies all four.}\label{tab:position}
\setlength{\tabcolsep}{2.3pt}
\begin{tabular}{lcccc}
\toprule
 & two & open & goal by & empty past \\
 & visits & vocab. & change rel. & locations \\
\midrule
RoomR~\cite{weihs2021roomr}            & yes & no  & no & no \\
EMQA~\cite{datta2022emqa}              & yes & no  & no & no (no nav.) \\
FindingDory~\cite{yadav2025findingdory}& yes & yes & no & no \\
3D-SMNet~\cite{cartillier2024smnet}    & yes & no  & partly & no (no nav.) \\
SuperMap~\cite{zhao2026supermap}       & yes & yes & no & no (history only) \\
OpenIN~\cite{openin2025}, OO-NAV~\cite{tang2025openobjectnav} & yes & yes & no & no \\
ObjectNav~\cite{batra2020objectnav}, VLN~\cite{anderson2018vln} & no & partly & no & no \\
OVMM~\cite{yenamandra2023homerobot}    & no  & yes & no & no \\
HM3D-OVON~\cite{yokoyama2024hm3dovon} & no  & yes & no & no \\
\textbf{OVMAN} & \textbf{yes} & \textbf{yes} & \textbf{yes} & \textbf{yes} \\
\bottomrule
\end{tabular}
\end{table}

\section{Methodology}\label{sec:method}

\subsection{Problem formulation}\label{sec:task}
\begin{figure*}[t]\centering
\includegraphics[width=\textwidth]{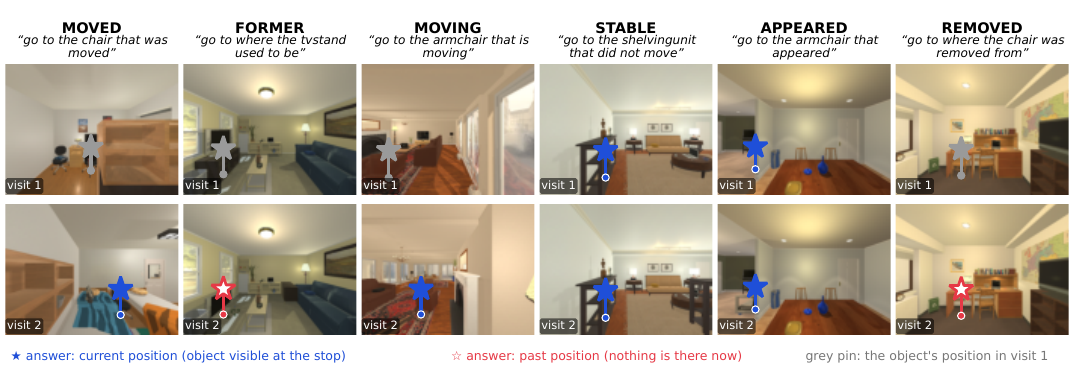}
\caption{The six relations on recorded episodes from the released set, one column each: the
agent's view on the familiarisation tour (top) and from the same tour after the change (bottom),
with the sentence it receives. The pin marks the answer, a floor-plane position: a filled star
where the answer is an object that must be visible at the stop, a hollow star where the answer
is a place the object has left. Frames are selected automatically by projecting the answer into
the recorded views, none by hand.}
\label{fig:relations}
\end{figure*}

Let $\mathcal{K}_v$ be the set of object instances present in visit $v$, and write the world state
as $\mathcal{W}_v = \{(k, c_k, x^{(v)}_k) : k \in \mathcal{K}_v\}$ with identity $k$, category
$c_k$ and floor-plane position $x^{(v)}_k \in \mathbb{R}^2$. An OVMAN episode is a triple
$(\mathcal{S}, \delta, q)$: a scene $\mathcal{S}$ in state $\mathcal{W}_1$, a scripted change
$\delta$ that produces $\mathcal{W}_2 = \delta(\mathcal{W}_1)$ while the agent is absent, and a
query $q = (c, \rho)$ naming a category and a change relation. Presence, not position, decides
membership, so we write $\mathcal{K}_{12} = \mathcal{K}_1 \cap \mathcal{K}_2$ for the instances
present in both visits,
$\mathcal{R}_{\text{cur}} = \{\textsc{moved}, \textsc{moving}, \textsc{stable}, \textsc{appeared}\}$
for the relations answered by a current position and
$\mathcal{R}_{\text{past}} = \{\textsc{former}, \textsc{removed}\}$ for those answered by a past
one. The agent observes each state
along the same fixed tour $\tau$ of $W$ waypoints at four headings, receives $q$ only after the
second tour, and must navigate to the answer. Writing
$\Delta_k = \lVert x^{(2)}_k - x^{(1)}_k \rVert$ for $k \in \mathcal{K}_{12}$, the six query
relations under this state model are the predicates on an instance $k$
\begin{equation}
\begin{aligned}
  \textsc{moved},\ \textsc{former} &:\ k \in \mathcal{K}_{12},\ \Delta_k \ge \Delta_{\min}, \\
  \textsc{stable} &:\ k \in \mathcal{K}_{12},\ \Delta_k < \tau, \\
  &\phantom{:\ } \exists\, k'\!\ne\! k:\ c_{k'}\!=\!c_k,\ \Delta_{k'}\!\ge\!\Delta_{\min}, \\[-1pt]
  \textsc{moving} &:\ x^{(2)}_k(t) \text{ changes with agent step } t, \\
  \textsc{appeared} &:\ k \in \mathcal{K}_2 \setminus \mathcal{K}_1, \\
  \textsc{removed} &:\ k \in \mathcal{K}_1 \setminus \mathcal{K}_2,
\end{aligned}
\label{eq:relations}
\end{equation}
These are six query relations under this state model, not six disjoint state transitions:
\textsc{moved} and \textsc{former} share a predicate and differ only in which endpoint is
returned. The model deliberately excludes articulated and appearance state, and each released
episode applies exactly one scripted change; several simultaneous changes are treated as a
diagnostic split, released alongside the main set. \textbf{Admissibility.} A query is released only
if its referent is unique,
\begin{equation}
  \bigl|\{\,k : c_k = c,\ \rho(k)\,\}\bigr| = 1,
  \label{eq:unique}
\end{equation}
and no instance of category $c$ other than the referent and the scripted-change object moves by
more than $\tau$ between the visits; the exception is what \textsc{stable} requires, since its
predicate is satisfied only when a same-category instance has moved. Both conditions are verified
on the recorded observations of every released episode. The answer is the position of
that instance,
\begin{equation}
  a_t(q) =
  \begin{cases}
    x^{(2)}_k & \rho \in \mathcal{R}_{\text{cur}} \setminus \{\textsc{moving}\},\\[-1pt]
    x^{(2)}_k(t) & \rho = \textsc{moving},\\[-1pt]
    x^{(1)}_k & \rho \in \mathcal{R}_{\text{past}},
  \end{cases}
  \label{eq:answer}
\end{equation}
which is fixed for every relation except \textsc{moving}, whose target advances with the agent's
own step count $t$. Two relations resolve to a \emph{past} position, a location where nothing is
(Fig.~\ref{fig:relations}), which is why the success test below is not uniform. Every query is
drawn from a category with $n \ge 2$ instances, so an agent that recognises the category but
not which of the $n$ instances the relation selects is at chance $1/n$.

Success follows the ObjectNav convention with one necessary departure. Let $p$ be the agent's
stopping position, $o_k$ the target object, $d_{\text{box}}(p, o_k)$ the distance from $p$ to
its axis-aligned bounding box, $\nu(a)$ the navigable cell nearest to $a$ in the spawn's
connected component, and $\mathrm{vis}(o_k)$ the predicate that the target's instance mask
covers at least $100$ pixels after a look-down sweep of $0^\circ$, $30^\circ$ and $60^\circ$.
Then
\begin{equation}
\begin{aligned}
  S_{\text{current}} &= \mathbb{1}\bigl[\min\bigl(d_{\text{box}}(p,o_k),\ \lVert p - \nu(a)\rVert\bigr) \le R\bigr] \wedge \mathrm{vis}(o_k),\\
  S_{\text{past}} &= \mathbb{1}\bigl[\min\bigl(\lVert p - a\rVert,\ \lVert p - \nu(a)\rVert\bigr) \le R\bigr],
\end{aligned}
  \label{eq:success}
\end{equation}
with $R = 1.0$\,m, where $a = a_{t_{\text{stop}}}(q)$ from \eqref{eq:answer} is the answer at the
step the agent calls stop, which matters only for \textsc{moving}. The box distance is needed because a wide object can place every navigable
cell beyond $R$ of its centre; the nearest navigable cell is the analogue of ObjectNav's
viewpoint sets, since an answer may lie on furniture or inside a footprint the agent cannot
occupy; and past-position answers carry no visibility term because nothing is there to see.
The test is applied identically to every agent, including the released baselines. We report
success rate and
\begin{equation}
  \mathrm{SPL} = S \cdot \frac{L^{*}}{\max(L^{*}, L)},
  \label{eq:spl}
\end{equation}
where $L^{*}$ is the geodesic length from the spawn to $\nu(a)$, the same reference point
\eqref{eq:success} scores against and $L$ the distance
travelled~\cite{anderson2018eval}. For \textsc{moving}, whose goal is not fixed, $L^{*}$ is
measured to the mover's recorded end point; SPL for moving targets has no agreed definition, so
success rate is primary there and this convention is stated rather than defended. Every arm is scored against a \emph{single-visit baseline}: the same perception with no retained
past, answering by picking uniformly among its grounded candidates for current-position relations
and, for past-position relations, by a Monte-Carlo estimate over uniform positions in the
reachable bounding box. It is a baseline rather than pure chance, because one visit still carries
within-visit evidence, especially for \textsc{moving}.

\subsection{Episode generation and certification}\label{sec:validity}
Algorithm~\ref{alg:gen} generates an episode. Geometric admissibility checks remove whole
classes of unsolvable candidates but are each outflanked by corner cases the simulator does not
expose, so acceptance is \emph{behavioural}: a candidate is emitted only if an oracle agent,
given the true answer, completes it with the same primitives and the same test
\eqref{eq:success} as the agents under evaluation: one generation-time feasibility run, then
three further certification runs. Repeated completion certifies that every released episode
\emph{admits} successful execution under the protocol; it does not make success certain on a
fresh run, because the \textsc{moving} mover is stochastic. Rejections are counted per query type
so the rule cannot bias the distribution unnoticed.

\begin{algorithm}[t]\footnotesize
\caption{Episode generation with certification}\label{alg:gen}
\begin{algorithmic}[1]
\Require scene $\mathcal{S}$, query type $\rho$, tour $\tau$ ($W$ farthest-point waypoints)
\State choose category $c$ with $\ge 2$ instances; choose target $k$ (and distractor $k'$ for \textsc{stable})
\State $\mathcal{O}_1 \gets$ observe $\tau$ in $\mathcal{W}_1$ \Comment{RGB-D, pose, instance masks $\ge 300$\,px}
\State apply $\delta$ to $k$, or to $k'$ for \textsc{stable}: displace by $\ge \Delta_{\min}$ with spawn clearance, or enable / disable it
\State \textbf{reject} unless $a(q)$ is reachable in the spawn's component and the spawn is mobile; for $\rho \in \mathcal{R}_{\text{cur}}$ also require $o_k$ visible from $\nu(a)$ \Comment{nothing is there to see for $\mathcal{R}_{\text{past}}$}
\State $\mathcal{O}_2 \gets$ observe $\tau$ in $\mathcal{W}_2$
\State \textbf{reject} unless every changed instance is recorded in the tour that defines its relation, and unless the referent is unique \eqref{eq:unique}
\State \textbf{reject} if any changed instance other than a \textsc{moving} mover drifted $> 0.25$\,m from its placement during the second tour
\For{$r = 1$ \textbf{to} $3$} \Comment{repeat-verified acceptance}
  \State \textbf{reject} unless the oracle agent, navigating to $a(q)$, satisfies \eqref{eq:success}
\EndFor
\State \Return $(\mathcal{O}_1, \mathcal{O}_2, q, a(q))$
\end{algorithmic}
\end{algorithm}

Lines 6--7 reject candidates whose recorded observations no longer describe the world the agents
are evaluated in, which happens when the simulator displaces an object after the change has been
applied.

\begin{figure*}[t]\centering
\includegraphics[width=\textwidth]{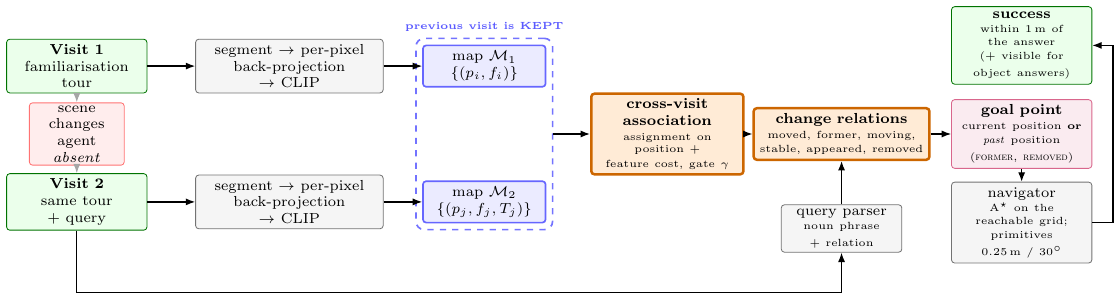}
\caption{The reference agent. Both per-visit maps are \emph{retained} (dashed box); the two
highlighted stages, cross-visit association and the change relations derived from it, are what a
single-state navigation stack or a current-state maintained map does not provide directly.}\label{fig:arch}
\end{figure*}

\begin{table}[t]\centering\scriptsize\renewcommand{\arraystretch}{0.92}
\caption{All constants of the benchmark and the reference agent, as read from the released code.}\label{tab:params}
\setlength{\tabcolsep}{3pt}
\begin{tabular}{llr}
\toprule
symbol & meaning & value \\
\midrule
$W$ & waypoints per tour (4 headings each) & 6 \\
$R$ & success radius & 1.0\,m \\
$\Delta_{\min}$ & minimum applied displacement & 1.5\,m \\
$\gamma$ & association gate & 1.25\,m \\
$\lambda$ & feature-disagreement scale & 2.0\,m \\
$\tau$ & moved / stable displacement threshold & 0.75\,m \\
$\tau_{\text{trk}}$ & moving track-spread threshold & 0.4\,m \\
$\delta_{\text{g}}$ & grounding band below best similarity & 0.03 \\
$\epsilon$ & instance clustering radius & 0.6\,m \\
--- & detector confidence / min.\ mask & 0.25 / 300\,px \\
--- & visibility: mask $\ge$ px, sweep & 100 / $0,30,60^\circ$ \\
--- & primitives: step, turn, heading tol. & 0.25\,m, $30^\circ$, $15^\circ$ \\
--- & action budget; stop within & 500; 0.60\,m \\
--- & blocked-edge expiry (\textsc{moving}); stall & 30; 80 actions \\
--- & validity radius; spawn clearance & 0.9\,m; 1.5\,m \\
--- & mover step per agent action & 0.10\,m \\
\bottomrule
\end{tabular}
\end{table}

\subsection{Reference agent}\label{sec:agent}
The reference agent (Fig.~\ref{fig:arch}; constants in Table~\ref{tab:params}) has five
stages. Its perception backbone follows the map construction of~\cite{ghosh2026vlmm}; the
cross-visit association and the change relations are new to this paper.

\emph{Perception.} For each tour frame with camera rotation $R \in SO(3)$, translation $t$,
intrinsics $K$ and depth $d_{uv}$, every pixel of a class-agnostic instance mask $m$
(YOLOv8m-seg) is back-projected and the mask is summarised by the component-wise median,
\begin{equation}
  X^{\text{w}}_{uv} = R\,d_{uv}\,K^{-1}[u,v,1]^{\top} + t, \quad
  \hat{X}_m = \operatorname{med}_{(u,v) \in m} X^{\text{w}}_{uv}.
  \label{eq:backproj}
\end{equation}
The median matters: a single ray at median depth is biased on non-convex masks and at depth
discontinuities and produced errors of up to $25.5$\,m in our ground-truth self-check over $262$ objects in the $20$ scenes. Each mask also contributes the CLIP embedding of its padded crop
(ViT-B/16, \texttt{laion2b\_s34b\_b88k}; the same encoder supplies the text embedding below). Detections over a tour are
clustered greedily at radius $\epsilon$ on the floor plane, giving for visit $v$ a map
$\mathcal{M}_v = \{(p_i, f_i, T_i)\}_{i=1}^{N_v}$: $p_i \in \mathbb{R}^2$ is the horizontal
projection of the cluster's median world points, $f_i$ the mean of the per-frame unit CLIP
embeddings renormalised to unit length, and $T_i$ the per-frame track.

\emph{Grounding.} With $f_q$ the CLIP text embedding of the noun phrase and
$s_i = \langle f_i, f_q\rangle$, the grounded set of a visit is
$G_v = \{\, i : s_i \ge \max_j s_j - \delta_{\text{g}} \,\}$, every instance within
$\delta_{\text{g}}$ of the best score rather than the single best, so that a near-tie does not
discard the correct instance.

\emph{Association.} $G_1$ and $G_2$ are matched by minimum-cost
assignment~\cite{kuhn1955hungarian} under
\begin{equation}
  C_{ij} = \lVert p_i - p_j \rVert_2 + \lambda\,\bigl(1 - \langle f_i, f_j \rangle\bigr),
  \label{eq:cost}
\end{equation}
where $\lambda$ converts feature disagreement into metres; the diagnostic arm that is given
ground-truth positions carries no features, so for it the cost reduces to pure distance. Non-assignment is
modelled by padding to a square matrix with one dummy per row and column at cost $D$ and
prohibitive cost elsewhere,
\begin{equation}
  \tilde{C} = \begin{bmatrix} C & D\,I + \infty(1 - I) \\ D\,I + \infty(1 - I) & 0 \end{bmatrix},
  \label{eq:pad}
\end{equation}
so leaving $i$ and $j$ unmatched costs $2D$ while pairing them costs $C_{ij}$ and frees both
dummies to pair at zero. A real--real edge with $C_{ij} \ge 2D$ is therefore never beneficial,
while edges with $C_{ij} < 2D$ are \emph{eligible} for the optimal assignment without being
guaranteed to win it, since sub-threshold edges still compete; setting $D = \gamma/2$ imposes the
admissibility gate $C_{ij} < \gamma$. The value of $\gamma$ is set on principle: under exact
positions any $\gamma < \Delta_{\min}$ prevents a displaced instance from re-matching its own
visit-1 detection; under real perception centroids carry error, so the choice is checked
empirically rather than assumed.

\emph{Decision.} Let $\mathcal{P}$ be the matched pairs and $U_1, U_2$ the unmatched sets of
$G_1, G_2$, and write $d(i, G) = \min_{j \in G} \lVert p_i - p_j \rVert$ for the distance from an
instance to its nearest counterpart in the other visit. Because $\gamma < \Delta_{\min}$,
non-assignment is the primary evidence and is consulted first:
\begin{equation}
\begin{aligned}
  \textsc{moved}, \textsc{appeared}:&\ \hat a = p_{j^\star},\ j^\star = \operatorname*{arg\,max}_{j \in U_2} d(j, G_1), \\[-1pt]
  \textsc{former}, \textsc{removed}:&\ \hat a = p_{i^\star},\ i^\star = \operatorname*{arg\,max}_{i \in U_1} d(i, G_2), \\[-1pt]
  \textsc{stable}:&\ \hat a = p_{j^\star},\ (i^\star\!, j^\star) = \operatorname*{arg\,min}_{\mathcal{P}} \lVert p_i\!-\!p_j\rVert,
\end{aligned}
\label{eq:rules}
\end{equation}
where \textsc{stable} answers only if that smallest displacement is below $\tau$, and any rule
whose candidate set is empty abstains rather than being evaluated on an empty argument. When $U_2$
(resp.\ $U_1$) is empty, \textsc{moved} and \textsc{former} fall back to the matched
pair of largest displacement provided it reaches $\tau$; \textsc{appeared} and \textsc{removed}
have no fallback and abstain, since an instance that matched across visits did not arrive or
leave. \textsc{moving} depends on the arm: with ground-truth positions it is the instance whose
within-visit track spread exceeds $\tau_{\text{trk}}$, whereas under real perception a moving
object fragments into several unmatched clusters, so the agent takes the latest-seen unmatched
cluster and answers with its last track point. Only \textsc{former} and \textsc{removed} return a goal
\emph{from} $\mathcal{M}_1$, which is why a map that keeps only the current state has no
mechanism for them. An agent that
cannot decide abstains and is scored as a failure.

\emph{Language and navigation.} A parser over the benchmark's vocabulary of surface forms
returns the noun phrase and the relation; a parse failure is a failure, never a default. Unless stated otherwise the agent is supplied with the category and the relation directly, and
the parser is evaluated separately on held-out surface forms. Navigation plans with A$^\star$ on the four-connected
grid of reachable positions recomputed after the change (the simulator's navigation mesh does not
carve moveable furniture); edges whose collision sweep fails are avoided, expiring after a fixed
horizon in \textsc{moving} episodes.

\section{Results and Discussion}\label{sec:results}
All episodes are generated and evaluated in AI2-THOR~\cite{kolve2017ai2thor} under
CloudRendering. All numbers are over the same 219 episodes with the test of \eqref{eq:success};
the agent is given the category and the relation unless stated otherwise. The single-visit rows are a
motion-aware map~\cite{ghosh2026vlmm} without a retained past, evaluated here as a component
rather than a competitor. Because every arm runs on the same episodes, differences between
arms are tested with exact paired McNemar tests on the discordant episodes, reported as wins to
losses; the exception is the single-visit arm, a stochastic pick rather than a realised policy, so
comparisons against it use the exact Poisson--binomial test of \S\ref{sec:solvable}. Everything runs on one laptop-class machine, an $8$-core Ryzen 7 7840HS with a single
RTX 4060 laptop GPU ($8$\,GB): mapping the recorded tours costs $1.9$\,s per episode, the static
gate then answers all $219$ episodes in $5$\,s, and one embodied pass over the four arms takes
$24$\,min against $30$--$34$\,min for CoW. We ask three questions.

\begin{table*}[t]\centering\small
\caption{Static gate: the predicted point must fall within $1.0$\,m of the answer. ``Grounding
handed over'' replaces a system's own language grounding with ground truth over \emph{its own}
map objects. DualMap is not run-to-run deterministic, so its rows are means over three runs;
ours are deterministic.}\label{tab:static}
\setlength{\tabcolsep}{3pt}\begin{tabular}{llcccccc c}
\toprule
system & memory & MOVED & FORMER & MOVING & STABLE & APPEARED & REMOVED & ALL \\
\midrule
ours (reference)      & single visit & 0.230 & 0.149 & 0.145 & 0.223 & 0.237 & 0.142 & 0.186 \\
DualMap~\cite{jiang2025dualmap} & maintained map & 0.202 & 0.158 & 0.139 & 0.198 & 0.133 & 0.183 & 0.167 \\
DualMap                & two visits   & 0.222 & 0.193 & 0.056 & 0.323 & 0.292 & 0.175 & 0.209 \\
ours (reference)       & two visits   & 0.545 & 0.447 & 0.361 & 0.281 & 0.375 & 0.475 & \best{0.416} \\
\midrule
\multicolumn{9}{l}{\emph{grounding handed over (diagnostic)}}\\
DualMap                & maintained map & 0.274 & 0.275 & 0.321 & 0.325 & 0.513 & 0.395 & 0.355 \\
DualMap                & two visits     & 0.424 & 0.237 & 0.352 & 0.292 & 0.517 & 0.358 & 0.365 \\
ours                   & single visit   & 0.369 & 0.328 & 0.271 & 0.358 & 0.525 & 0.495 & 0.396 \\
ours                   & two visits     & 0.485 & 0.342 & 0.500 & 0.656 & 0.525 & 0.400 & \best{0.479} \\
\midrule
\multicolumn{9}{l}{\emph{positions handed over (diagnostic)}}\\
ours                   & single visit   & 0.389 & 0.079 & 0.371 & 0.500 & 0.556 & 0.072 & 0.321 \\
ours                   & two visits     & 1.000 & 1.000 & 0.972 & 1.000 & 1.000 & 1.000 & \best{0.995} \\
\bottomrule
\end{tabular}
\end{table*}

\begin{table}[t]\centering\footnotesize
\caption{Embodied evaluation: the agent navigates with the primitives of Table~\ref{tab:params}. CoW runs from its released code~\cite{gadre2023cows}.}\label{tab:embodied}
\setlength{\tabcolsep}{2.2pt}\begin{tabular}{lcccccccc}
\toprule
agent & MVD & FMR & MVG & STB & APP & RMV & SR & SPL\\
\midrule
CoW\,\cite{gadre2023cows} & 0.121 & 0.026 & 0.194 & 0.219 & 0.150 & 0.050 & 0.123 & --- \\
ours, 1 visit & 0.424 & 0.132 & 0.389 & 0.375 & 0.325 & 0.100 & 0.283 & 0.269 \\
ours, 2 visits & 0.576 & 0.421 & 0.472 & 0.406 & 0.425 & 0.425 & 0.452 & 0.440 \\
ours, oracle pos & 1.000 & 0.974 & 0.972 & 1.000 & 1.000 & 1.000 & 0.991 & 0.982 \\
oracle check & 1.000 & 1.000 & 0.972 & 1.000 & 1.000 & 1.000 & 0.995 & 0.990 \\
\bottomrule
\end{tabular}
\end{table}

\begin{table}[t]\centering\footnotesize
\caption{How the two-visit agent fails (static gate). ``Wrong instance'': within $R$ of another
instance of the category. The first five rows partition the $219$ episodes; ``lost in
navigation'' is a \emph{subset} of \textup{success} --- static successes lost when executed.}\label{tab:taxonomy}
\setlength{\tabcolsep}{3pt}\begin{tabular}{lccccccc}
\toprule
outcome & MVD & FMR & MVG & STB & APP & RMV & ALL \\
\midrule
success & 18 & 17 & 13 & 9 & 15 & 19 & 91 \\
abstain (no candidate) & 9 & 10 & 12 & 2 & 17 & 17 & 67 \\
elsewhere ($>2$\,m) & 4 & 10 & 5 & 17 & 5 & 3 & 44 \\
near miss (1--2\,m) & 1 & 0 & 3 & 3 & 2 & 0 & 9 \\
wrong instance & 1 & 1 & 3 & 1 & 1 & 1 & 8 \\
\midrule
episodes & 33 & 38 & 36 & 32 & 40 & 40 & 219 \\
\midrule
\multicolumn{8}{l}{\emph{subset of \textup{success}, not a sixth outcome:}}\\
lost in navigation & 0 & 1 & 1 & 0 & 0 & 2 & 4 \\
\bottomrule
\end{tabular}
\end{table}

\begin{figure}[t]\centering
\includegraphics[width=\columnwidth]{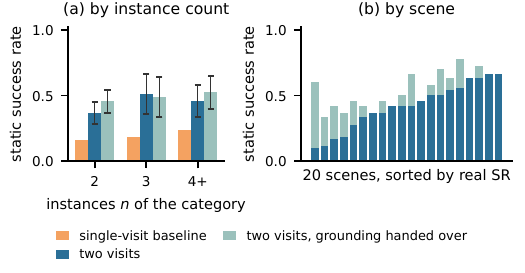}
\caption{Static success of the two-visit agent (a) by the number of instances of the queried
category (123, 37 and 59 episodes), with 95\% intervals,
against the single-visit baseline and the same agent with grounding handed over; (b) by
scene.}\label{fig:difficulty}
\end{figure}

\subsection{Is the task solvable, and does memory matter?}\label{sec:solvable}
Yes, and only for the relations built on the past. Table~\ref{tab:static}
gates predictions statically and Table~\ref{tab:embodied} executes them. With oracle positions
the two-visit rule reaches $0.991$ embodied, so the episodes are solvable and the decision
rules are not the obstacle. Under real perception the two-visit agent succeeds on $91$ of $219$
episodes where the single-visit arm is expected to succeed on $40.8$; because that arm is a
stochastic pick among grounded candidates rather than a realised policy, the admissible exact test
is the Poisson--binomial tail of its own per-episode probabilities, which gives $p<10^{-22}$
overall. Per relation, and correcting for the six comparisons, the margin holds for
\textsc{removed}, \textsc{former} ($p<10^{-8}$), \textsc{moved} ($p<10^{-4}$) and
\textsc{moving} ($p<10^{-3}$), is marginal for \textsc{appeared} ($p=0.032$) and is
\emph{not} significant for \textsc{stable} ($p=0.243$). When navigating the margin is 59 to 22
($p<10^{-4}$); per relation only \textsc{removed} survives the same correction ($p=0.014$), with
\textsc{former} at $p=0.064$. Memory is largest where the answer is a past position, but it is not
confined there: it also helps on \textsc{moved} and \textsc{moving}, where the second visit must
still be told which instance the relation selects. It is \textsc{stable} --- the one relation whose
answer a single visit can in principle read off directly --- where memory buys nothing. The 95\% intervals are $0.228$--$0.346$ for one visit,
$0.388$--$0.518$ for two and $0.967$--$0.997$ with oracle positions; per-relation
cells hold 32--40 episodes and are read as trends. Navigation itself is not the difficulty:
when the two-visit agent succeeds its path is near-optimal, SPL given success $0.973$, and
the embodied rate exceeds the static one (12 to 4) because the ObjectNav stop test accepts a
prediction that lands $1$--$2.6$\,m from the answer point yet within $R$ of the object's box, which
happened on 12 episodes, mostly \textsc{moving} and \textsc{stable}. Relations differ in how
they fail: \textsc{stable} is the weakest statically because its rule commits to a matched pair
and commits to the wrong one when the target is missing from the grounded set; \textsc{moving}
fragments a moving object into several clusters; \textsc{appeared} and \textsc{removed} abstain
most often, having no fallback by design (Table~\ref{tab:taxonomy}).

\subsection{Do the existing paradigms fail, and why?}
CoW~\cite{gadre2023cows} reaches the change-defined answer in $12.3\%$ of episodes and
scores $0.026$ on \textsc{former} ($0.035$ over three seeds): a single-visit agent has almost no
way to represent a location where the object no longer is. On its own task, reaching any instance
of the queried category, the same runs score $0.333$; the gap to $0.123$ is the anti-shortcut rule, and
the two-visit agent beats CoW on 83 episodes to 11 ($p<10^{-14}$). CoW runs through its released
runner with policy, thresholds, prompts and embodiment unmodified; the four infrastructure
changes it needed are released as a patch.

\begin{figure}[t]\centering
\includegraphics[width=\columnwidth]{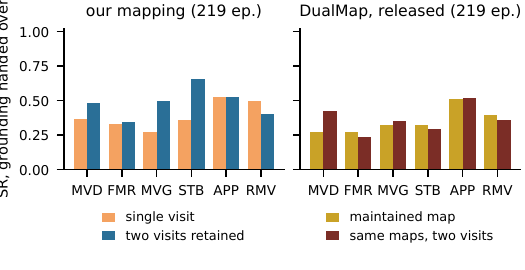}
\caption{Two visits against one current-state reading, per relation, with grounding handed over:
our mapping (left), the released DualMap (right). The gain is on the relations that still leave an
object to find; on \textsc{former} and \textsc{removed} it is negligible or reversed.}\label{fig:memory}
\end{figure}

Fig.~\ref{fig:memory} isolates the architectural claim. With grounding handed over, DualMap's
maintained map answers \textsc{former} at $0.275$ and \textsc{removed} at $0.395$, while the same
maps read as two visits answer them at $0.237$ and $0.358$, and overall $0.355$ against $0.365$;
on our mapping the totals move from $0.396$ to $0.479$. Reading the same maps as two visits rather
than one maintained map therefore does \emph{not} recover the past-position relations once the
diagnostic is made honest: the overall difference is $0.010$, and on the two past-position
relations the maintained map is ahead. The architectural claim is not supported by this arm. On
one of the 20 scenes the released system degenerates to about $7$ retained objects per map against
$43$--$89$ elsewhere; its detector crashed on the empty first tour frame of that scene, and an
18-line guard, released as a patch, leaves its policy, thresholds and prompts unchanged.
Per-scene results are released alongside the aggregate.

\begin{figure}[t]\centering
\includegraphics[width=\columnwidth]{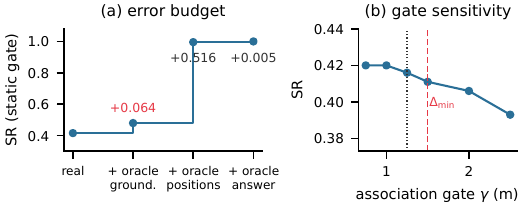}
\caption{(a) Error budget on the static gate; each rung hands over one error source. (b) Static
success against the association gate $\gamma$: the span over $[0.75, 2.5]$\,m is $0.027$, and the
value used lies below $\Delta_{\min}$ by construction.}\label{fig:diag}
\end{figure}

\subsection{Where does the remaining difficulty lie?}
Fig.~\ref{fig:diag}a reads the rungs of Table~\ref{tab:static} as a budget: from $0.416$, oracle
grounding lifts the agent to $0.479$, oracle positions to $0.995$, and the given answer to
$1.000$ by definition. Grounding costs only $0.064$ of the gap and the handover is not significant
(49 episodes gained to 35 lost, $p=0.16$); the mapping itself --- localisation and cross-visit
association --- costs $0.516$, and answer selection $0.005$. The difficulty is
therefore not in naming the object but in carrying an instance across two visits. Navigation is almost free of the
budget: with positions handed over, executing the plan costs $0.005$ ($0.995$ static against
$0.991$ embodied).

Table~\ref{tab:taxonomy} locates the failures. The agent rarely guesses wrong: 8 of 219
predictions land on another instance. It abstains (67) or predicts a position more than $2$\,m
away (44) --- the conservative branch of \eqref{eq:rules} when CLIP ranks a distractor above the
target --- and navigation loses 4 of the 91 static successes. Difficulty is not combinatorial
(Fig.~\ref{fig:difficulty}): success is $0.366$, $0.514$ and $0.458$ for two, three and four or
more same-category instances, but runs from $0.100$ to $0.667$ across scenes, and grounding
handed over lifts every scene to at least $0.500$.

\emph{Multi-change diagnostic split.} We also release the 170 multi-change episodes that pass the
same certification, in which two or three instances of the queried category change under
different relations. With grounding handed over the agent reaches $0.676$ on it, and the relation
that fails is \textsc{appeared} ($0.185$): the new instance must be told apart from the moved one,
and appearance cues do not close it: fitting CLIP crop
features, a hue--saturation histogram and the back-projected extent into the association cost, in the
four variants we release (two of which are no-ops, the fitted weights placing all mass on the CLIP
term), moves the split from $0.676$ to $0.671$. The cues carry weak signal --- the
best same-category AUROC is $0.588$ in-sample on the only split with enough pairs to fit, and
chance ($0.500$) on the held-out split --- but it does not
convert into task success, which is why an instance identity these maps do not carry, rather than a
better appearance descriptor, is the prerequisite.

\emph{Validity checks.} The agent parses a held-out phrasing for both the object and the relation
at $0.941$, and success is unchanged whether the relation is supplied or inferred ($0.406$ and
$0.406$), since the only confusions (\textsc{removed} read as \textsc{former}) resolve to the same
answer location. Across three seeds --- which reseed the reference agent's own stochastic choices, not the
simulator --- the two-visit agent scores
$0.452$, $0.461$ and $0.461$, and the single-visit agent $0.283$, $0.256$ and $0.247$; the embodied
margin between them is therefore $0.169$--$0.215$, wider than either arm alone suggests; the
variation is the simulator's, since the static gate is identical under all three seeds, which is why
the paired static comparison over the same $219$ episodes is the stronger evidence. CoW is
re-run under the same three seeds, through the seeding its own runner already provides, so
Table~\ref{tab:embodied} does not compare against a single run: it reaches $0.119$--$0.128$ across the three
(mean $0.123$) and $0.026$--$0.053$ on \textsc{former}. DualMap consumes fixed recorded sequences yet is not run-to-run
deterministic either, so it too is repeated three times; with grounding handed over the
two-visit reading does not beat the maintained map under any pairing, the worst pairing being
$-0.029$ overall and $-0.135$ on \textsc{former}. Nor does the
conclusion rest on where the admissibility thresholds were set: re-deriving admissibility over $256$
defensible settings of the drift tolerance, $\tau$, $\Delta_{\min}$ and each filter, the static
memory margin stays within $0.224$--$0.247$ and never changes sign. The embodied oracle is $0.995$ rather than exactly $1.000$ because the
\textsc{moving} mover is stochastic (Section~\ref{sec:validity}).

\section{Conclusion}
OVMAN poses a navigation goal whose referent is a relation between two world states, including
the case where the answer is a place an object has left, and certifies every episode solvable by
repeated oracle execution. The two released systems closest to it fail in structurally different
ways: a single-visit navigator has no mechanism to stop where nothing is, and a self-maintaining
map answers past-tense queries poorly, though reading the same maps as two visits does not by
itself recover them either --- the gain from two visits is on the relations that still leave an
object to find.

Two limitations define the scope. Episodes contain a single change in one simulator's indoor
scenes; the multi-change split shows that several simultaneous changes require an instance
identity across visits that these maps do not carry, so multi-view shape or learned
re-identification is the prerequisite for solving it, and we leave that to future work. The
reference agent is not a proposed method; its error budget marks where stronger agents, in
particular ones that learn the cross-visit association, should gain.

\end{document}